\documentclass[10pt]{article}

\usepackage[affil-it]{authblk}

\usepackage{cite}
\usepackage{amsmath,amssymb,amsfonts}
\usepackage{algorithmic}
\usepackage{graphicx}
\usepackage{textcomp}
\usepackage{subcaption}
\usepackage{caption}
\usepackage{algorithm}

\providecommand{\keywords}[1]{\textbf{Keywords:} #1}

\begin{document}

\title{A Neural Template Matching Method to Detect Knee Joint Areas}
\author{Juha Tiirola%
\thanks{Email: \texttt{juha.a.tiirola@gmail.com}}}
\affil{Research Unit of Mathematical Sciences, University of Oulu, Finland}
\date{\today}

\maketitle

\begin{abstract}
In this paper, new methods are considered to detect knee joint areas in  bilateral PA fixed flexion knee X-ray images. The methods are of template matching type where the distance criterion is based on the negative normalized cross-correlation. The manual annotations are made on only one side of a single bilateral image when the templates are selected. The best matching patch search is formulated as an unconstrained continuous domain minimization problem. For the minimization problem different optimization methods are considered. The main method of the paper is a trainable optimizer where the method is taught to take zoomed and possibly rotated patches from its input images which look like the template. In the experiments, we compare the minimum values found by different optimization methods. We also look at some test images to examine the correspondence between the minimum value and how well the knee area is localized. It seems that making annotations only to a single image enables to detect knee joint areas quite precisely.
\end{abstract}


\section{Introduction}
Knee osteoarthritis (OA) refers to a disease where the knee articular cartilage is degreded. The major features to assess the disease are joint space narrowing, formation of osteophytes and sclerosis.  The classical X-ray imaging based grading system to knee osteoarthritis is the KL grading system, proposed by Kellgren and Lawrence in \cite{Kellgren_Lawrence} which gives an integer score from $0$ to $4$ to each knee image. In this system, the  higher the score, the more severe the disease. The KL grading system is based on the features found in the knee joint area. Thus if one wants to exclude the influence of the background on the KL classification, a natural first step is to localize the knee joint areas from the bilateral image. In \cite{Antony2020}, an overview is given about the disease and the problems of localizing the joint area and the automatic assesment of OA severity. In \cite{SHAMIR20091307}, template matching is used to automatically detect knee joint areas. They use the Euclidean distance as the closeness criterion and $20$ template images. In \cite{Antony2020}, the template matching method of \cite{SHAMIR20091307} is also discussed and they state that the  detections by the method may depend much on the choice of the template images. Other approaches also exist, in \cite{Pierson} the whole bilateral X-ray image was used as an input for learning to measure the severity of osteoarthritis. In \cite{Lewis95fastnormalized}, the fast computation of normalized cross-correlation is considered. If template matching is formulated by  comparing the template to each possible patch in the input image in a sliding window way, it is  not so easy to incorporate other prior information, for instance that the left and right knee joint areas are nearly of equal size, and still get fast computation methods. In contrast, if the template matching is formulated as a continuous domain minimization problem, it is easy to add regularizing terms to the template matching energy and one can use the methods of gradient based optimization instead of brute-force based approaches to find the optimal patch.

In \cite{Antony2016QuantifyingRK}, Sobel horizontal image gradients are used  as the input features for detecting knee joints. In training, the knee joint areas are manually selected in each training image. This leads to positive and negative samples depending on whether the center of the labeled Region of Interest (ROI) is inside the sample. Using a linear SVM classifier they learn to classify negative and positive samples. In the test phase, the maximal score is taken as the center of the detected ROI and a fixed size neighbourhood is taken as the detected ROI.  In \cite{Antony2017AutomaticDO},  binary images with masks specifying the manually annotated knee joint areas are used as the training labels. Then a fully convolutional network is learnt  to map the input image to a soft pixel-based classification map. In the test phase, the estimated ROIs are obtained from the soft output by contour detection. In \cite{tiulpin2017novel}, a two-step method is proposed to the knee joint area detection where first knee joint area proposals are generated based on the limb anatomy. Then these proposals are scored based on their Histogram of oriented gradients features. Training of the scoring step is based on the SVM classifier with positive and negative samples depending on the amount of overlap with the manually selected bounding boxes. In \cite{CHEN201984} and \cite{Wang_Bi}, YOLOv2 method of \cite{redmon_farhardi} is finetuned in a manually annotated data set to find knee joint areas. In \cite{GuLi}, Yolo v3-tiny of \cite{Adarsh} is trained to detect knee joint areas.

In \cite{NIPS2015_33ceb07b}, the spatial transformer module was introduced to warp feature maps in a differentiable way. The module can be inserted into neural network architectures without a need for extra labeling. In the method, given a parametric family of transformations, the localization net analyses its input feature map and outputs a parameter vector. The transformation corresponding to the parameter vector is then used to warp the input feature map. If the set of  affine transformations is used as the parametric family, the spatial transformer module enables to take zoomed and rotated patches from its input feature map in a differentiable way.

Gradient descent is the classical optimization approach also in the training of neural networks. For instance in \cite{Andrychowicz}, also the design of optimization algorithms is cast as a learning problem.  They proposed to learn a neural network optimizer when different but related minimization problems are  available as the training data, and they show that the learnt optimizer can generalize to unseen, but nevertheless training-related minimization problems. One example problem considered in \cite{Andrychowicz} is the style transfer problem \cite{Gatys}. In the problem, given a style patch, an input image is transformed in a structure preserving way into a same style image as the given style patch. The same style image is obtained as the solution of a minimization problem. If the input image is changed, a different but related minimization problem is obtained.

In this paper, we consider a neural network based method to the knee joint area localization problem. As the training data, we use $500$ bilateral knee X-ray images and manual annotations are made only on a single X-ray image. The annotation includes selecting a bounding box which forms the template image. In addition, we select two smaller boxes inside the template image at the left and right sides of the knee joint borders. We show the manual annotation done in Fig.\ \ref{fig:template_T}. Our method uses the sampling mechanism of the spatial transformer to obtain a patch from the input image in a differentiable way. The spatial transformer module warps the input image such that the warped output looks like the template patch. The corresponding transformation parameter vector is found by solving a minimization problem where the energy expression to be minimized is based on the negative, normalized cross-correlation between the extracted patches and the template image plus some regularization. The minimization problem has $8$ parameters. Since for each training image there is a minimization problem, similar to \cite{Andrychowicz}, we learn the optimizer.  The architecture of the optimizer is a Resnet18 backbone followed by two additional 1-1 convolutional layers followed by a regression network which consists of a three layer fully connected neural network. 

For comparison, we also consider to solve each minimization problem separately. The minimization problem is highly non-convex so the outputs of the gradient based optimization algorithms depend heavily on the initialization. To minimize the energy for a given bilateral image, we run a gradient based optimization for many different initializations. The initializations are based on a grid. First, the grid has a few scales. For a given scale, many translations are tested for the initial center point of the candidate optimal patch. When the gradient based optimization is run, we record all loss values. Finally, we select the parameter corresponding to the smallest loss encountered. This method is time consuming since for a given bilateral X-ray image, depending on the grid there may be thousands of optimization problems to be solved.

\section{Pre-processing}
\begin{figure}[ht!]
\centering
\includegraphics[width=7.5cm]{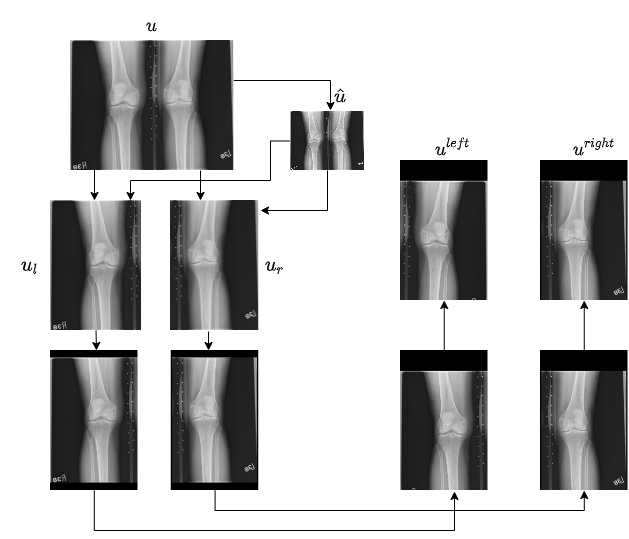}
\caption{The deterministic splitting and random augmentation steps. The image $u$ is split first into somewhat overlapping halves $u_l$ and $u_r$. Then zeros are added to make the ratio of the height over width of each half a constant. This deterministic step is followed by random translation, the horizontal flip  of the left half and resizing both halves to $800\times500$  pixels.}
\label{split_aug}
\end{figure}

The input images are bilateral X-ray knee images as $u$ in Fig.\  \ref{split_aug}. The goal is to detect the knee joint areas from the input image. 
 After the preprocessing we make the problem simplifies somewhat in the sense that there is only one knee joint area from each half to be found. An overview of the split process and augmentation is shown in Fig.\ \ref{split_aug}.

We use a down-sampled version of $u$ to roughly divide $u$ into vertically symmetric halves. If $h_0$ is the horizontal coordinate of the center point of the downsampled image $\hat{u}$, then for each $a\in\{h_0-0.125 width(\hat{u}),h_0+0.125 width(\hat{u})\}$ we do the following: let $w:=\min\{a,\text{width}(\hat{u})-a\}$. Compute the average Euclidean distance between $\hat{u}[:,a-w:a]$ and $\text{horizontal}\_\text{flip}(\hat{u}[:,a:a+w])$. Select the $a$ giving the smallest average distance as the location where the split is done. Then widen the non-overlapping halves by the factor $1.1$ to get the slightly overlapping halves $u_l$ and $u_r$. Let $w_l$ and $w_r$ be the width of $u_l$ and $u_r$, respectively. If $w_l>w_r$, then we widen $u_r$ by adding zeros to its both sides by equal amounts so that the augmented $u_r$ has width $w_l$. Otherwise, we widen $u_l$. As a result both $u_l$ and $u_r$ are of the same size $h\times w$. If $\frac{h}{w}>1.6$, we widen both $u_l$ and $u_r$ by adding zeros to both sides such that $\frac{h}{w}$ becomes $1.6$. If $\frac{h}{w}<1.6$, we add zeros both on top and bottom parts. The whole process has been so far deterministic. Next we add a small random translation in the training step. After this we resize both halves to size $800\times500$ pixels and we flip the left part horizontally. We get images which we call $u^{left}$ and $u^{right}$.

\section{Parametrization}\label{sect:parametrization}
The extraction of the knee joint areas from $u^{left}$ and $u^{right}$ is based on the use of the sampling mechanism of the spatial transformer. Given a parametric family of transformations the spatial transformer has a localization net which takes a feature map as an input and outputs transformation parameters. The transformation corresponding to the parameters is used to produce sampling points in the input feature map domain. Then a sampling mechanism is used to produce a warped feature map by interpolating the input feature map at the sampling points. Here we would like to place a possibly rotated rectangle inside each input image such that the rectangle defines the joint area. A parameter vector specifies the scale, location and rotation of the rectangle. Since the area must be inside the input image, the parametrization is constrained. The parameter vector $\pmb{\theta}^s$ determines the corresponding transformation $A(\pmb{\theta}^s)$ which is used to warp a regular grid to obtain the sampling points. Each half of Fig.\ \ref{split_aug} has its own transformation parameter vector. The regular grid of the input $u^s$ lies in $[-1,1]^2$ and the sampling points lie inside the warped square $\{\,A(\pmb{\theta}^s)\begin{pmatrix}\pmb{x}\\1\end{pmatrix} \,|\, \pmb{x}\in[-1,1]^2 \,\}$. We want the warped square to be inside $[-1,1]^2$. We denote by $u^s(\pmb{\theta}^s)$ the corresponding warped image obtained by applying the sampler of the spatial transformer to interpolate $u^s$ at the sampling points. 
The template matching tries to find a suitable $\pmb{\theta}^s$ such that the distance between $u^s(\pmb{\theta}^s)$ and the template is as small as possible.

Let $f:=\text{height}(\text{template})/\text{width}(\text{template})$. We assume that the template is such that $f\geq1$. A warped image has its own regular normalized coordinates $[-1,1]^2$. This square is multiplied coordinatewise by the scale vector $(s_1,s_2):=(\pmb{\theta}^s_1,\pmb{\theta}^s_1/f)$, possibly rotated slightly by $\pmb{\theta}^s_4$ radians, translated by a vector $(\pmb{\theta}^s_2,\pmb{\theta}^s_3)$, and the resulting transformed coordinate square must lie inside the $[-1,1]^2$ coordinate square of the input half image. Thus, $\pmb{\theta}^s$ is constrained and required to be meaningfully interpretable. 

Instead of searching the transformation parameters directly as the solution of a constrained minimization problem, we consider an unconstrained optimization problem where an arbitrary, unconstrained parameter vector $\pmb{v}^s\in\mathbb{R}^4$ is mapped to an interpretable transformation vector $\pmb{\theta}^s\in\mathbb{R}^4$. 

Let $\pmb{v}^s\in\mathbb{R}^4$ be arbitrary. First we consider how the scale$\in(0,1)$ is computed from $\pmb{v}^s_1$. In Fig.\ \ref{split_aug} we see a typical input halves $u^{left}$ and $u^{right}$ from which we try to extract   patches which look like the template patch $\pmb{T}$ of Fig.\ \ref{fig:template_T}. The sizes of the extracted patches cannot be arbitrarily small, instead the width of the patch is typically like a third of the width of the halves. Thus we map the domain $\mathbb{R}$ of $\pmb{v}^s_1$ into a meaningful interval subset of $(0,1)$. We set $\alpha_0=0.15$, $\beta_0=0.8$, and  
\begin{equation}\label{eq:scale_choice}
\pmb{\theta}^s_1:=\alpha_0+\beta_0\left(\frac{1}{2}\left(1+\tanh(\pmb{v}^s_1)\right)\right).
\end{equation} 
We see that $\pmb{\theta}^s_1\in[\alpha_0,\alpha_0+\beta]\subset[0,1]$. 

Next we come to the choice of the translation parameters. 
For each scale, the possible locations for the center of the zoomed patch are constrained such that the extracted patch lies fully inside the input image. If $\pmb{\theta}^s_2$ gives the horizontal center location of the rectangle of width $2 s_1$, then we get constraints $-1\leq \pmb{\theta}^s_2-s_1$ and $\pmb{\theta}^s_2+s_1\leq1$, and thus $-1+s_1\leq\pmb{\theta}^s_2\leq 1-s_1$. If we define  
\[
a_{13}:=\frac{1}{2}(1+\tanh(\pmb{v}^s_2)) \text{ and } a_{23}:=\frac{1}{2}(1+\tanh(\pmb{v}^s_3))
\] 
and then set 
\begin{equation}\label{eq:tr_choice}
\begin{split}
\pmb{\theta}^s_2&:=-1+s_1+a_{13}\cdot2(1-s_1)=(1-s_1)(-1+2 a_{13}), \\
\pmb{\theta}^s_3&:=-1+s_2+a_{23}\cdot2(1-s_2)=(1-s_2)(-1+2 a_{23})
\end{split}
\end{equation}
we get the above constraints satisfied.
 
In addition, we add a possibility for small rotation. We set
\[
\pmb{\theta}^s_4:=0.13 \tanh(\pmb{v}^s_4).
\]
The constraint for the location was for a non-rotated rectangle so in theory a part of a rotated rectangle might be outside $[-1,1]^2$ but this was negligible in the experiments.

We have described how to transform an unconstrained $\pmb{v}^s\in\mathbb{R}^4$ to an interpretable transformation vector $\pmb{\theta}^s\in\mathbb{R}^4$.
We denote by 
\begin{equation}\label{eq:D_def}
D:\mathbb{R}^4\to\mathbb{R}^4
\end{equation} 
the above map $\pmb{v}^s\mapsto\pmb{\theta}^s$.

If $\pmb{\theta}^s=D(\pmb{v}^s)$, the transformation matrix corresponding to $\pmb{v}^s$ is given by
\begin{equation}\label{eq:tr_matrix_A}
A(\pmb{\theta}^s):=\begin{bmatrix}
\pmb{\theta}^s_1 \cos(\pmb{\theta}^s_4) & -\pmb{\theta}^s_1 \sin(\pmb{\theta}^s_4) & \pmb{\theta}^s_2 \\
(\pmb{\theta}^s_1/f) \sin(\pmb{\theta}^s_4) & (\pmb{\theta}^s_1/f) \cos(\pmb{\theta}^s_4) & \pmb{\theta}^s_3
\end{bmatrix}.
\end{equation}
When the above matrix is applied to a vector $\pmb{x}\in[-1,1]^2$, it is understood that $\pmb{x}$ is augmented first by adding $1$ as the third dimension, so the result of applying the matrix to the vector is $A(\pmb{\theta}^s)\begin{pmatrix}\pmb{x}\\1\end{pmatrix}\in\mathbb{R}^2$, see \cite{NIPS2015_33ceb07b}.

\section{Loss function}\label{sect:loss}
\begin{figure}[ht!]
\centering
\includegraphics[width=6.5cm]{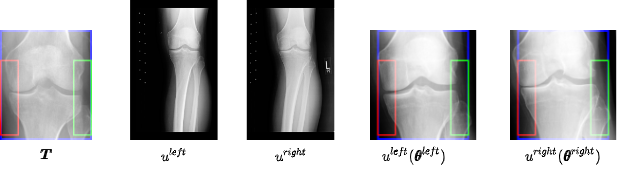}
\caption{A template $\pmb{T}$ and its two sub-templates shown by red and green boxes. The images $u^s(\pmb{\theta}^s)$ at the regular grid points $\pmb{x}$ are obtained by interpolating $u^s$ at the points $A(\pmb{\theta}^s)[\pmb{x},1]^T$.}
\label{fig:template_T}
\end{figure}

We manually select a template $\pmb{T}$ and its two sub-templates from a training image, see Fig.\ \ref{fig:template_T}. If we have a loss function $L$, which measures the distance between warped half images and the template, then a way to find the warping transformations is to solve the unconstrained minimization problem
\[
\min_{(\pmb{v}^{left},\pmb{v}^{right})\in\mathbb{R}^8} L(u^{left}(D(\pmb{v}^{left})),\pmb{T})+L(u^{right}(D(\pmb{v}^{right})),\pmb{T}).
\]
Next we consider how to choose the loss function. 
Let $\mathbf{u}$ and $\mathbf{v}$ be two images. Let $\tilde{\mathbf{u}}$ and $\tilde{\mathbf{v}}$ be the same images with their means removed. Define a global registration cost using the negative normalized cross-correlation as 
\[
c^{gl}(u,v):=1-\frac{\tilde{\mathbf{u}}\cdot\tilde{\mathbf{v}}}{||\tilde{\mathbf{u}}|| \, ||\tilde{\mathbf{v}}||}.
\]
In addition to the global window indicated by the blue rectangle at the boundary in Fig.\ \ref{fig:template_T}, we also consider costs over two sub-windows whose fixed locations are indicated by the red and green windows in Fig.\ \ref{fig:template_T}. We define sub-costs
\[
c^r(u,v):=1-\frac{\widetilde{\mathbf{u}|_{rw}}\cdot\widetilde{\mathbf{v}|_{rw}}}{||\widetilde{\mathbf{u}|_{rw}}|| \, ||\widetilde{\mathbf{v}|_{rw}}||},
~
c^g(u,v):=1-\frac{\widetilde{\mathbf{u}|_{gw}}\cdot\widetilde{\mathbf{v}|_{gw}}}{||\widetilde{\mathbf{u}|_{gw}}|| \, ||\widetilde{\mathbf{v}|_{gw}}||}
\]
where for instance $u|_{rw}$ denotes $u$ restricted to the interior of the red bounding box. We add the global cost and sub-costs and define
\[
\tilde{c}(u,v):=\frac{1}{2}\left(c^{gl}(u,v)+\max\{c^r(u,v),c^g(u,v)\}\right).
\]
Since there are also image negatives in the training data, we take this into account and define the loss function by
\[
L(u,v):=\min\{\tilde{c}(u,v),\tilde{c}(-u,v)\}.
\]
As visible also in Fig.\ \ref{fig:template_T}, the knee joint areas between halves are quite equal sized and in addition the vertical coordinates of the center of the knee joint areas are near each other. Since $\pmb{\theta}^s_1$ gives the size and $\pmb{\theta}^s_3$ the vertical coordinate of the center of the extracted patch, we would like to have $\pmb{\theta}_1^{left}\approx\pmb{\theta}_1^{right}$ and $\pmb{\theta}_3^{left}\approx\pmb{\theta}_3^{right}.$ If we have a bilateral image $u$ and $\pmb{v}:=(\pmb{v}^{left},\pmb{v}^{right})\in\mathbb{R}^8$, we define a regularized cost as 
\begin{equation}\label{eq:f}
\begin{split}
f^{\pmb{T}}(u,\pmb{v}):=&\mathcal{L}^{left}+\mathcal{L}^{right}+\mathcal{L}^{reg} \\
:=&L(u^{left}(D(\pmb{v}^{left})),\pmb{T})+L(u^{right}(D(\pmb{v}^{right})),\pmb{T}) \\
+&||D(\pmb{v}^{left})_1-D(\pmb{v}^{right})_1||_2^2+||D(\pmb{v}^{left})_3-D(\pmb{v}^{right})_3||_2^2 \\
=:& L(u^{left}(D(\pmb{v}^{left})),\pmb{T})+L(u^{right}(D(\pmb{v}^{right})),\pmb{T}) \\
+&L^{reg}(D(\pmb{v}^{left}),D(\pmb{v}^{right})) \\
=:& L(u^{left},u^{right},\pmb{v},\pmb{T}).
\end{split}
\end{equation}
If for a fixed $u$, standard gradient based optimization is used to find a candidate for the minimizer $\pmb{v}$ of $f^{\pmb{T}}(u,\pmb{v})$, the vector found depends on the initial point. One can try to minimize $\pmb{v}\mapsto f^{\pmb{T}}(u,\pmb{v})$ by doing optimization with several different initial points, where the initial points are constructed from a grid, and finally taking the argument which gives the smallest recorded cost. This is highly time consuming since depending on the density of the grid there may be thousands of optimization problems for each $u$. We consider a possible choice of the grid in Appendix.

\section{Learning the minimizer}
\begin{figure*}[ht!]
\centering
\includegraphics[width=12.0cm]{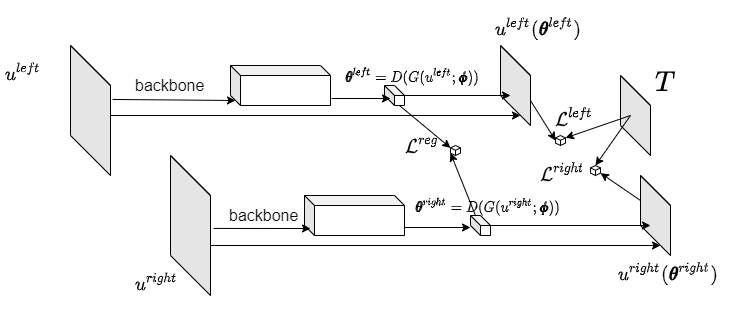}
\caption{In the siamese network, the backbone consists of the convolutional part of the Resnet18 network \cite{he_kaiming} followed by two additional $1\times1$ convolutional layers with a ReLU nonlinearity between the additional layers. If $N$ is the batch size, the shape of the Resnet part output changes as $(N,512,7,7)\overset{\text{additional layer}}{\longrightarrow} (N,128,7,7)\overset{\text{ReLU}}{\longrightarrow}(N,128,7,7)\overset{\text{additional layer}}{\longrightarrow} (N,32,7,7)\overset{\text{ReLU}}{\longrightarrow}(N,32,7,7)\overset{\text{Normalize, $dim=1$}}{\longrightarrow}(N,32,7,7)$.  The last tensor is reshaped to $(N,32\cdot7\cdot7)$ and goes through a feed-forward network, which has three fully connected layers: $(N,1568)\overset{\text{linear}}{\longrightarrow}(N,512)\overset{\text{ReLU}}{\longrightarrow}(N,512)\overset{\text{linear}}{\longrightarrow}(N,32)\overset{\text{ReLU}}{\longrightarrow}(N,32)\overset{\text{linear}}{\longrightarrow}(N,4)\overset{6(-\frac{1}{2}+\sigma(\cdot))}{\longrightarrow}(N,4).$ The last output corresponding to input $u^s$ is denoted by $\pmb{v}^s$, so $\pmb{v}^s=G(u^s;\pmb{\phi})$. Also, $\pmb{\theta}^s:=D(\pmb{v}^s).$
}
\label{fig:architecture}
\end{figure*}

As in \cite{Andrychowicz} we consider how to learn the minimizer of \eqref{eq:f}. The learnt minimizer $G(\cdot;\pmb{\phi})$ takes $u^s$ as input and outputs $\pmb{v}^s:=G(u^s;\pmb{\phi})$, $s\in\{right,left\}.$ The architecture of $G$ shown in Fig.\ \ref{fig:architecture} is of Siamese type so the weights of the left and right branches are shared. The fully connected part has the last layer $x\mapsto 6(-0.5+\sigma(x))$ so that when the $\tanh$ nonlinearity is applied to the output of the fully connected part  later in $D$, the output of $\tanh$ has its range near $(-1,1)$. In Fig.\  \ref{fig:architecture}, $D\circ G(\cdot;\pmb{\phi})$ is called the Localisation net in \cite{NIPS2015_33ceb07b}.
\begin{figure*}[ht!]
\centering
\includegraphics[width=12.0cm]{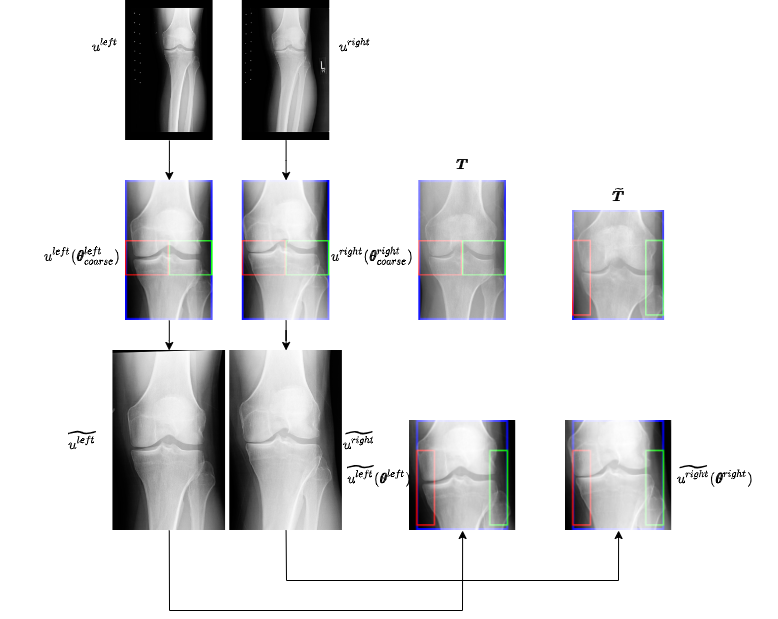}
\caption{Outline of the two-phase algorithm. Template $\pmb{T}$ and its sub-template $\tilde{\pmb{T}}$ are extracted from the same X-ray image.}
\label{fig:two_phases}
\end{figure*}

We denote $G(u;\pmb{\phi}):=(G(u^{left};\pmb{\phi}),G(u^{right};\pmb{\phi}))$. We learn the parameters $\pmb{\phi}$ by minimizing the function
\begin{equation}\label{eq:L_phi}
\mathcal{L}(\pmb{\phi}):=E_u[c_u f^{\pmb{T}}(u,G(u;\pmb{\phi}))]
\end{equation}
where $E$ denotes the expectation and $u$ is sampled from the space of bilateral X-ray knee images. We add a scaling factor $c_u$, $c_u>0$, in the expectation since the value of $\min_{\pmb{v}} f^{\pmb{T}}(u,\pmb{v})$ is not constant with respect to $u$ and as a result,  in training, if $B$ is a minibatch, then in $1/|B|\sum_{i\in B} f^{\pmb{T}}(u_i;G(u_i;\pmb{\phi}))$ a possible much larger than others summand can dominate and training tries to make this large value as small as possible. To treat minimization problems more equally, we add the scaling $c_u$. A possible choice of $c_u$ could be $c_u=1/\min_{\pmb{v}} f^{\pmb{T}}(u,\pmb{v})$. But, as was argued earlier, this is slow to compute, so we use an approximation to it.   

We do the search of the optimal patch in two phases. We outline this in Fig.\  \ref{fig:two_phases} and Algorithm~\ref{alg:two}. First, we learn to search the template $\pmb{T}$ from the input images $u^{left}$, $u^{right}$. This phase produces a network $G(\cdot;\pmb{\phi}_{coarse})$. The input images for the second phase are slightly enlargened $u^{left}(D(G(u^{left};\pmb{\phi}_{coarse})))$ and $u^{right}(D(G(u^{right};\pmb{\phi}_{coarse})))$. We denote the enlargened images by $\widetilde{u^{left}}$ and $\widetilde{u^{right}}.$ From these images the second network $G(\cdot;\pmb{\phi})$ is taught to extract patches that look like the template $\tilde{\pmb{T}}$ of Fig.\ \ref{fig:two_phases}.

\begin{algorithm}
\caption{A Phase Training}\label{alg:cap}
\begin{algorithmic}
\REQUIRE $\text{Training data }\{u_i\}_{i=1}^N, \text{ number of outer iterations }$ 
\STATE $T, \text{ number of epochs } M, \text{ template } \pmb{T}$
\STATE $t \gets 0$
\WHILE{$t \neq T$}
\IF{$t==0:$}
\STATE Set $c_{u_i}=1$ for all $u_i$
\ELSE
\STATE For each $u_i:=(u^{left}_i,u^{right}_i)$ do sharpening: use $\pmb{v}_0:=(G(u^{left}_i;\pmb{\phi}^{t}),G(u^{right}_i;\pmb{\phi}^{t}))$ as an  initial point for the minimization of $f^{\pmb{T}}(u_i,\pmb{v}).$ Use the Adam optimizer with $300$ iterations. Record all values of $f^{\pmb{T}}(u_i,\pmb{v}_j)$ for $j=0,1,\dotsc,300$, and set $c_{u_i}=1/\min_j f^{\pmb{T}}(u_i,\pmb{v}_j)$.  
\ENDIF
\STATE $t \gets t+1$
\STATE $\text{Update } \pmb{\phi} \text{ in } \eqref{eq:L_phi} \text{ by running } M \text{ epochs with minibatch}$ 
\STATE $\text{size } 8 \text{ over the training data.}$ We get $G(\cdot;\pmb{\phi}^t).$
\ENDWHILE \\
\RETURN $G(\cdot;\pmb{\phi}^T)$
\end{algorithmic}
\end{algorithm}

\begin{algorithm}
\caption{Two Phases Training}\label{alg:two}
\begin{algorithmic}
\STATE \textbf{Phase 1}: Using $\{u_i:=(u^{left}_i$, $u^{right}_i)\}_{i=1}^{500}$ and the template $\pmb{T}$ from Fig.\ \ref{fig:two_phases}, apply Algorithm \ref{alg:cap} with $T=3$, $M=15$, to obtain $G(\cdot;\pmb{\phi}^{coarse})$.
\STATE \textbf{Phase 2}: Enlarge the images $u_i^{left}(D(G(u_i^{left};\pmb{\phi}^{coarse})))$ and $u_i^{right}(D(G(u_i^{right};\pmb{\phi}^{coarse})))$ by $1.2$ for all $i=1,\dotsc,500,$ and rename the enlarged pairs as $u_i$, select the template $\widetilde{\pmb{T}}$ from Fig.\  \ref{fig:two_phases} as $\pmb{T}$, $T=3$, $M=15$, and apply Algorithm \ref{alg:cap} to obtain $G(\cdot;\pmb{\phi})$.
\end{algorithmic}
\end{algorithm}
We train the neural model using $500$ bilateral Osteoarthritis Initiative (OAI) images. The images are from the baseline i.e.\ from the first visit. We minimize the energy using the Adam optimizer with batch size $8$, the  learning rate $1e-5$ for the Resnet18 backbone parameters and $1e-4$ for the rest of the parameters. We use PyTorch and we initialize the Resnet18 parameters using its pre-trained version. 

\section{Experimental results}
We compare four methods in the numerical experiments. In the experiments, we  first compare minimums found by different methods by examining the magnitudes of the non-regularized loss values $\mathcal{L}^{left}(P^{left},\tilde{\pmb{T}})$ and $\mathcal{L}^{right}(P^{right},\tilde{\pmb{T}})$ in \eqref{eq:f}, and also the sum values $\mathcal{L}^{left}(P^{left},\tilde{\pmb{T}})+\mathcal{L}^{right}(P^{right},\tilde{\pmb{T}})$  where $P^{left}$ and $P^{right}$ are the detected areas. We also visually compare whether there is a correspondence between how small the loss value is and how good the obtained knee joint area localization is. We do not consider for instance the classical intersection over union values, since the continuous domain methods we consider also allow for rotations, and it is quite difficult and laborious to annotate possibly rotated bounding boxes, the step we wanted to avoid in the first place.	 

The baseline template matching method we consider is based on Python's scikit-image's  match$\_$template function which is based on \cite{Lewis95fastnormalized}. We use the method to compare in a sliding window manner the template $\tilde{\pmb{T}}$ in Fig.\ \ref{fig:two_phases} to five differently scaled versions of the input image and we extract the patches corresponding to the smallest left plus right loss value. The second method we consider is the neural network method in Algorithm~\ref{alg:two}. The third method is the second method followed by the sharpening step described inside Algorithm~\ref{alg:cap}, so the trained neural network method is used to get a good initialization point which is followed by $300$ iterations with the Adam optimizer and the patch corresponding to the smallest value recorded is taken as the output patch of the third method. The fourth method is described at the end of Section~\ref{sect:loss} and in Appendix. 

In the first experiment we consider how small the minimums found by different methods are on average between different methods. We consider images from the OAI dataset and the Multicenter Osteoarthritis Study (MOST) dataset. The OAI images are from the first visit but different from the training data. In Tables~\ref{table:cross_corr_dataset_OAI} and \ref{table:cross_corr_dataset_MOST} we consider the averages of the values $\mathcal{L}^{left}(P^{left},\tilde{\pmb{T}})+\mathcal{L}^{right}(P^{right},\tilde{\pmb{T}})$ by three different methods. It seems that in the OAI dataset the baseline template matching method does not have dense enough scale grid or since the method is non-regularized and does not allow for rotations, the method does not find as small minimums as the neural method or  its sharpening step. In the MOST dataset the neural method does not directly find as small minimums as in the OAI dataset, but the method produces such  good initial points that the sharpening step again finds much smaller minimums. 

\begin{figure*}
\center
\begin{tabular}{c|c}
\begin{subfigure}{0.1\textwidth}
\includegraphics[width=0.95\textwidth]{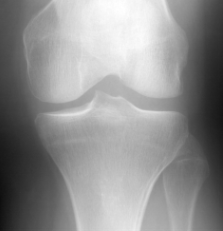}
\vspace{-1.4\baselineskip}
\caption*{$0.1212$}
\end{subfigure}
\begin{subfigure}{0.1\textwidth}
\includegraphics[width=0.95\textwidth]{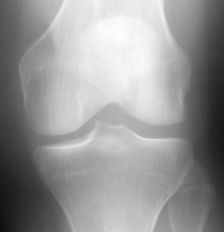}
\vspace{-1.4\baselineskip}
\caption*{$0.0604$}
\end{subfigure}
\begin{subfigure}{0.1\textwidth}
\includegraphics[width=0.95\textwidth]{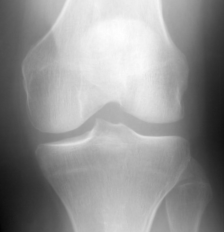}
\vspace{-1.4\baselineskip}
\caption*{$0.0563$}
\end{subfigure}
\begin{subfigure}{0.1\textwidth}
\includegraphics[width=0.95\textwidth]{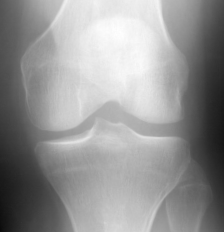}
\vspace{-1.4\baselineskip}
\caption*{$0.056$}
\end{subfigure}
&
\begin{subfigure}{0.1\textwidth}
\includegraphics[width=0.95\textwidth]{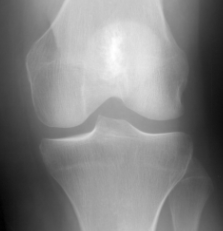}
\vspace{-1.4\baselineskip}
\caption*{$0.0717$}
\end{subfigure}
\begin{subfigure}{0.1\textwidth}
\includegraphics[width=0.95\textwidth]{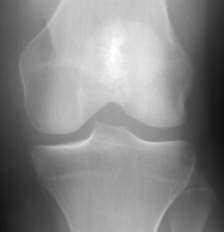}
\vspace{-1.4\baselineskip}
\caption*{$0.0686$}
\end{subfigure}
\begin{subfigure}{0.1\textwidth}
\includegraphics[width=0.95\textwidth]{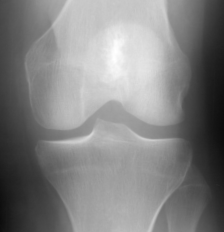}
\vspace{-1.4\baselineskip}
\caption*{$0.0622$}
\end{subfigure}
\begin{subfigure}{0.1\textwidth}
\includegraphics[width=0.95\textwidth]{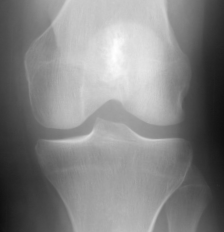}
\vspace{-1.4\baselineskip}
\caption*{$0.0626$}
\end{subfigure}
\\[5ex]
\begin{subfigure}{0.1\textwidth}
\includegraphics[width=0.95\textwidth]{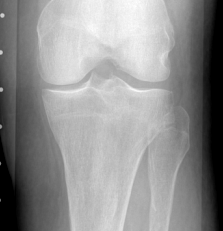}
\vspace{-1.4\baselineskip}
\caption*{$0.0892$}
\end{subfigure}
\begin{subfigure}{0.1\textwidth}
\includegraphics[width=0.95\textwidth]{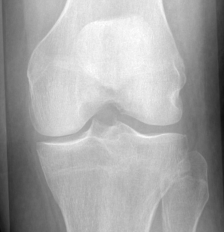}
\vspace{-1.4\baselineskip}
\caption*{$0.078$}
\end{subfigure}
\begin{subfigure}{0.1\textwidth}
\includegraphics[width=0.95\textwidth]{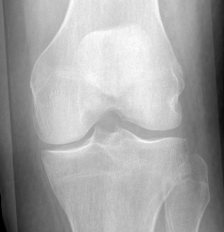}
\vspace{-1.4\baselineskip}
\caption*{$0.0658$}
\end{subfigure}
\begin{subfigure}{0.1\textwidth}
\includegraphics[width=0.95\textwidth]{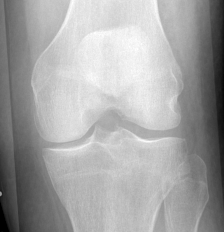}
\vspace{-1.4\baselineskip}
\caption*{$0.0654$}
\end{subfigure}
&
\begin{subfigure}{0.1\textwidth}
\includegraphics[width=0.95\textwidth]{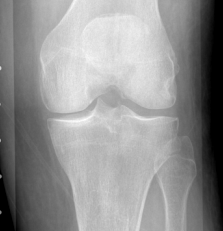}
\vspace{-1.4\baselineskip}
\caption*{$0.1083$}
\end{subfigure}
\begin{subfigure}{0.1\textwidth}
\includegraphics[width=0.95\textwidth]{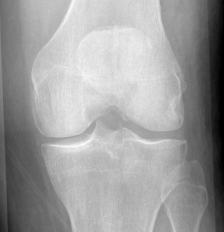}
\vspace{-1.4\baselineskip}
\caption*{$0.0766$}
\end{subfigure}
\begin{subfigure}{0.1\textwidth}
\includegraphics[width=0.95\textwidth]{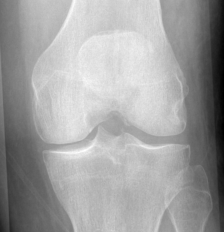}
\vspace{-1.4\baselineskip}
\caption*{$0.0715$}
\end{subfigure}
\begin{subfigure}{0.1\textwidth}
\includegraphics[width=0.95\textwidth]{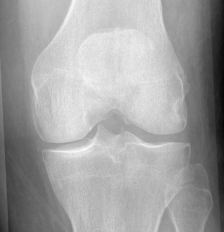}
\vspace{-1.4\baselineskip}
\caption*{$0.0714$}
\end{subfigure}
\\[5ex]
\begin{subfigure}{0.1\textwidth}
\includegraphics[width=0.95\textwidth]{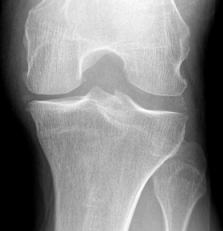}
\vspace{-1.4\baselineskip}
\caption*{$0.099$}
\end{subfigure}
\begin{subfigure}{0.1\textwidth}
\includegraphics[width=0.95\textwidth]{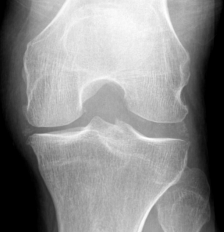}
\vspace{-1.4\baselineskip}
\caption*{$0.0663$}
\end{subfigure}
\begin{subfigure}{0.1\textwidth}
\includegraphics[width=0.95\textwidth]{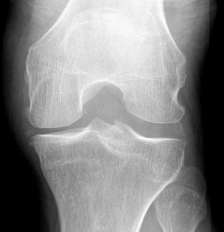}
\vspace{-1.4\baselineskip}
\caption*{$0.062$}
\end{subfigure}
\begin{subfigure}{0.1\textwidth}
\includegraphics[width=0.95\textwidth]{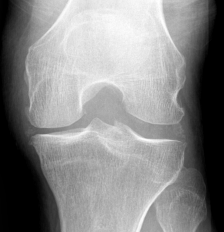}
\vspace{-1.4\baselineskip}
\caption*{$0.0628$}
\end{subfigure}
&
\begin{subfigure}{0.1\textwidth}
\includegraphics[width=0.95\textwidth]{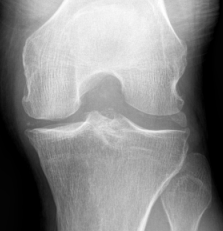}
\vspace{-1.4\baselineskip}
\caption*{$0.1064$}
\end{subfigure}
\begin{subfigure}{0.1\textwidth}
\includegraphics[width=0.95\textwidth]{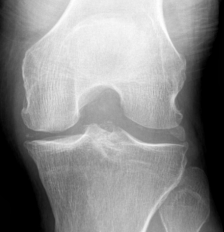}
\vspace{-1.4\baselineskip}
\caption*{$0.0949$}
\end{subfigure}
\begin{subfigure}{0.1\textwidth}
\includegraphics[width=0.95\textwidth]{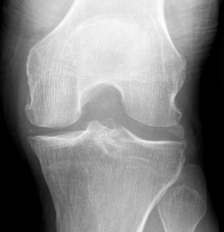}
\vspace{-1.4\baselineskip}
\caption*{$0.0822$}
\end{subfigure}
\begin{subfigure}{0.1\textwidth}
\includegraphics[width=0.95\textwidth]{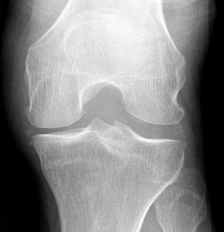}
\vspace{-1.4\baselineskip}
\caption*{$0.0628$}
\end{subfigure}
\\[5ex]
\begin{subfigure}{0.1\textwidth}
\includegraphics[width=0.95\textwidth]{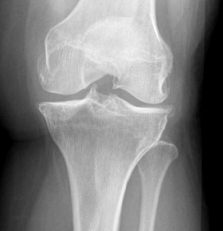}
\vspace{-1.4\baselineskip}
\caption*{$0.2258$}
\end{subfigure}
\begin{subfigure}{0.1\textwidth}
\includegraphics[width=0.95\textwidth]{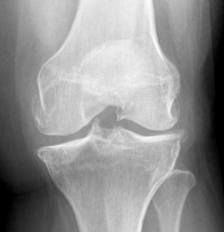}
\vspace{-1.4\baselineskip}
\caption*{$0.106$}
\end{subfigure}
\begin{subfigure}{0.1\textwidth}
\includegraphics[width=0.95\textwidth]{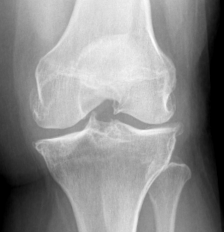}
\vspace{-1.4\baselineskip}
\caption*{$0.0873$}
\end{subfigure}
\begin{subfigure}{0.1\textwidth}
\includegraphics[width=0.95\textwidth]{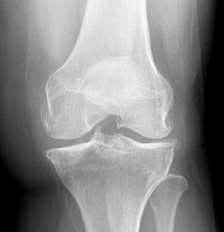}
\vspace{-1.4\baselineskip}
\caption*{$0.0803$}
\end{subfigure}
&
\begin{subfigure}{0.1\textwidth}
\includegraphics[width=0.95\textwidth]{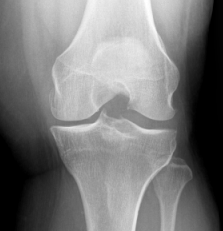}
\vspace{-1.4\baselineskip}
\caption*{$0.1474$}
\end{subfigure}
\begin{subfigure}{0.1\textwidth}
\includegraphics[width=0.95\textwidth]{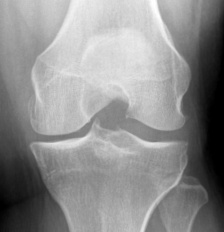}
\vspace{-1.4\baselineskip}
\caption*{$0.1047$}
\end{subfigure}
\begin{subfigure}{0.1\textwidth}
\includegraphics[width=0.95\textwidth]{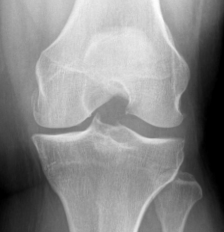}
\vspace{-1.4\baselineskip}
\caption*{$0.0905$}
\end{subfigure}
\begin{subfigure}{0.1\textwidth}
\includegraphics[width=0.95\textwidth]{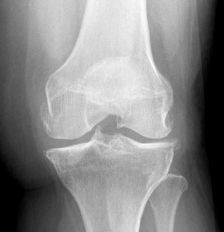}
\vspace{-1.4\baselineskip}
\caption*{$0.0803$}
\end{subfigure}
\\[5ex]
\begin{subfigure}{0.1\textwidth}
\includegraphics[width=0.95\textwidth]{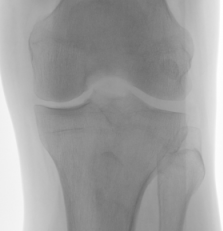}
\vspace{-1.4\baselineskip}
\caption*{$0.0872$}
\end{subfigure}
\begin{subfigure}{0.1\textwidth}
\includegraphics[width=0.95\textwidth]{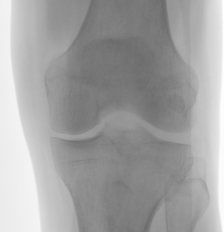}
\vspace{-1.4\baselineskip}
\caption*{$0.0917$}
\end{subfigure}
\begin{subfigure}{0.1\textwidth}
\includegraphics[width=0.95\textwidth]{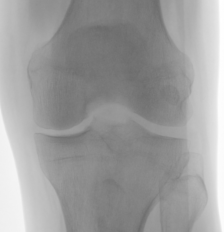}
\vspace{-1.4\baselineskip}
\caption*{$0.0362$}
\end{subfigure}
\begin{subfigure}{0.1\textwidth}
\includegraphics[width=0.95\textwidth]{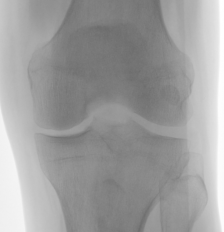}
\vspace{-1.4\baselineskip}
\caption*{$0.0366$}
\end{subfigure}
&
\begin{subfigure}{0.1\textwidth}
\includegraphics[width=0.95\textwidth]{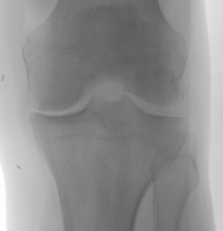}
\vspace{-1.4\baselineskip}
\caption*{$0.0985$}
\end{subfigure}
\begin{subfigure}{0.1\textwidth}
\includegraphics[width=0.95\textwidth]{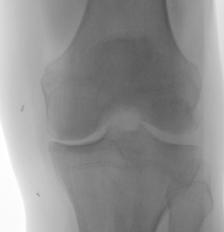}
\vspace{-1.4\baselineskip}
\caption*{$0.1649$}
\end{subfigure}
\begin{subfigure}{0.1\textwidth}
\includegraphics[width=0.95\textwidth]{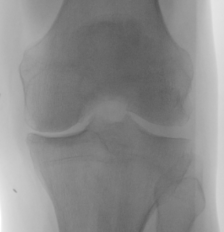}
\vspace{-1.4\baselineskip}
\caption*{$0.0667$}
\end{subfigure}
\begin{subfigure}{0.1\textwidth}
\includegraphics[width=0.95\textwidth]{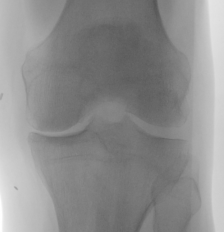}
\vspace{-1.4\baselineskip}
\caption*{$0.065$}
\end{subfigure}
\\[5ex]
\begin{subfigure}{0.1\textwidth}
\includegraphics[width=0.95\textwidth]{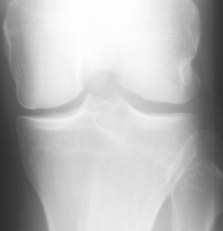}
\vspace{-1.4\baselineskip}
\caption*{$0.0941$}
\end{subfigure}
\begin{subfigure}{0.1\textwidth}
\includegraphics[width=0.95\textwidth]{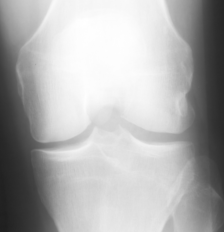}
\vspace{-1.4\baselineskip}
\caption*{$0.0587$}
\end{subfigure}
\begin{subfigure}{0.1\textwidth}
\includegraphics[width=0.95\textwidth]{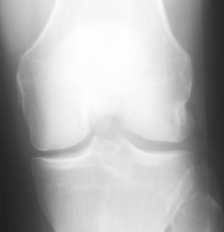}
\vspace{-1.4\baselineskip}
\caption*{$0.0528$}
\end{subfigure}
\begin{subfigure}{0.1\textwidth}
\includegraphics[width=0.95\textwidth]{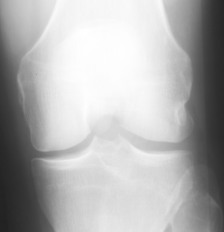}
\vspace{-1.4\baselineskip}
\caption*{$0.0525$}
\end{subfigure}
&
\begin{subfigure}{0.1\textwidth}
\includegraphics[width=0.95\textwidth]{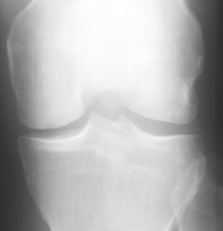}
\vspace{-1.4\baselineskip}
\caption*{$0.0753$}
\end{subfigure}
\begin{subfigure}{0.1\textwidth}
\includegraphics[width=0.95\textwidth]{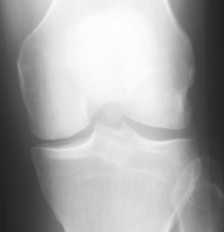}
\vspace{-1.4\baselineskip}
\caption*{$0.0723$}
\end{subfigure}
\begin{subfigure}{0.1\textwidth}
\includegraphics[width=0.95\textwidth]{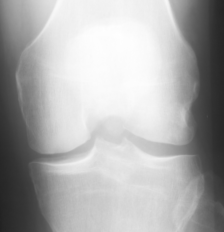}
\vspace{-1.4\baselineskip}
\caption*{$0.0586$}
\end{subfigure}
\begin{subfigure}{0.1\textwidth}
\includegraphics[width=0.95\textwidth]{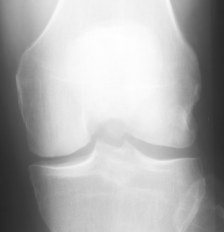}
\vspace{-1.4\baselineskip}
\caption*{$0.0573$}
\end{subfigure}
\\[5ex]
\begin{subfigure}{0.1\textwidth}
\includegraphics[width=0.95\textwidth]{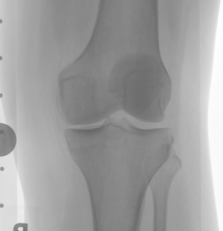}
\vspace{-1.4\baselineskip}
\caption*{$0.1164$}
\end{subfigure}
\begin{subfigure}{0.1\textwidth}
\includegraphics[width=0.95\textwidth]{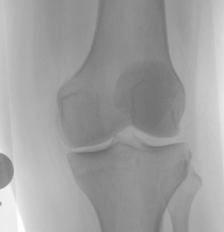}
\vspace{-1.4\baselineskip}
\caption*{$0.1393$}
\end{subfigure}
\begin{subfigure}{0.1\textwidth}
\includegraphics[width=0.95\textwidth]{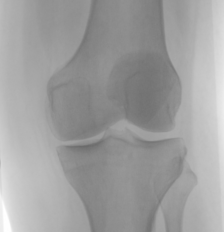}
\vspace{-1.4\baselineskip}
\caption*{$0.1142$}
\end{subfigure}
\begin{subfigure}{0.1\textwidth}
\includegraphics[width=0.95\textwidth]{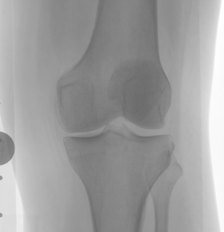}
\vspace{-1.4\baselineskip}
\caption*{$0.0966$}
\end{subfigure}
&
\begin{subfigure}{0.1\textwidth}
\includegraphics[width=0.95\textwidth]{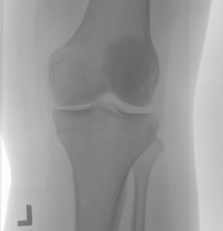}
\vspace{-1.4\baselineskip}
\caption*{$0.3015$}
\end{subfigure}
\begin{subfigure}{0.1\textwidth}
\includegraphics[width=0.95\textwidth]{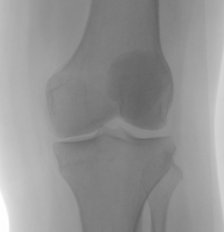}
\vspace{-1.4\baselineskip}
\caption*{$0.1196$}
\end{subfigure}
\begin{subfigure}{0.1\textwidth}
\includegraphics[width=0.95\textwidth]{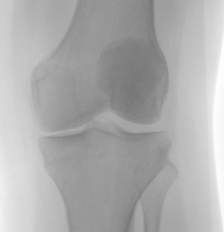}
\vspace{-1.4\baselineskip}
\caption*{$0.0939$}
\end{subfigure}
\begin{subfigure}{0.1\textwidth}
\includegraphics[width=0.95\textwidth]{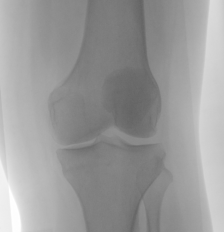}
\vspace{-1.4\baselineskip}
\caption*{$0.0964$}
\end{subfigure}
\\[5ex]
\begin{subfigure}{0.1\textwidth}
\includegraphics[width=0.95\textwidth]{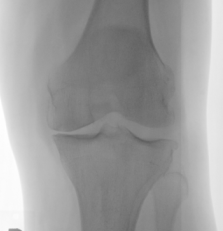}
\vspace{-1.4\baselineskip}
\caption*{$0.08$}
\end{subfigure}
\begin{subfigure}{0.1\textwidth}
\includegraphics[width=0.95\textwidth]{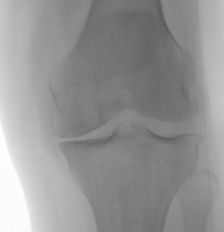}
\vspace{-1.4\baselineskip}
\caption*{$0.232$}
\end{subfigure}
\begin{subfigure}{0.1\textwidth}
\includegraphics[width=0.95\textwidth]{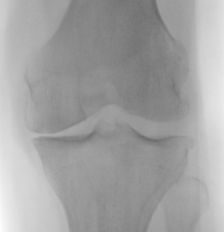}
\vspace{-1.4\baselineskip}
\caption*{$0.0878$}
\end{subfigure}
\begin{subfigure}{0.1\textwidth}
\includegraphics[width=0.95\textwidth]{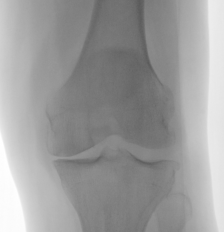}
\vspace{-1.4\baselineskip}
\caption*{$0.0727$}
\end{subfigure}
&
\begin{subfigure}{0.1\textwidth}
\includegraphics[width=0.95\textwidth]{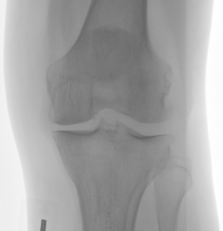}
\vspace{-1.4\baselineskip}
\caption*{$0.0863$}
\end{subfigure}
\begin{subfigure}{0.1\textwidth}
\includegraphics[width=0.95\textwidth]{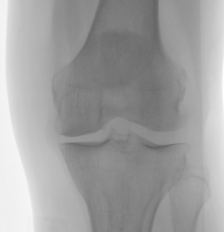}
\vspace{-1.4\baselineskip}
\caption*{$0.1122$}
\end{subfigure}
\begin{subfigure}{0.1\textwidth}
\includegraphics[width=0.95\textwidth]{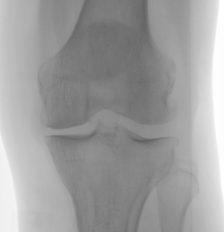}
\vspace{-1.4\baselineskip}
\caption*{$0.0729$}
\end{subfigure}
\begin{subfigure}{0.1\textwidth}
\includegraphics[width=0.95\textwidth]{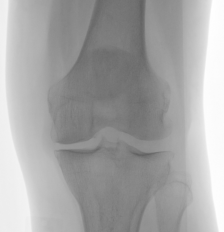}
\vspace{-1.4\baselineskip}
\caption*{$0.0669$}
\end{subfigure}
\end{tabular}
\caption{The detections and the corresponding one-sided loss values obtained by four different methods on eight test images. The vertical bar separates the sides. The first and fifth column: the template matching method. The second and sixth column: the neural method. The third and seventh column: the neural method followed by the sharpening step. The fourth and eighth column: the grid initialized search algorithm.}
\label{fig:eight_test}
\end{figure*}

In Fig.\ \ref{fig:eight_test} we show some typical test detections by different methods and the corresponding one-sided values  $\mathcal{L}^{left}(P^{left},\tilde{\pmb{T}})$ and $\mathcal{L}^{right}(P^{right},\tilde{\pmb{T}})$. From the experiments we made, it seems that in the baseline template matching method, there is more variability in the patches found and there is more vertical translations between the sides compared to other methods. In contrast, the neural method and its sharpening step produce detections which are more uniform. This is again due to regularization incorporated  into the continuous domain methods. It seems that the very time consuming fourth method based on the grid initialized search does not necessarily find smaller loss values than the neural network method followed by the sharpening step. Also, in the neural network method, the parameters describing the knee joint areas are based on the last convolutional layer features of the Resnet18 architecture which are not necessarily so sensitive to small variations in the input image. This might explain why the fourth method does not produce as uniform detections as the neural method.

\begin{table}[!htbp]
\small
\centering
\caption{The sides combined average values over $400$ test images in the OAI dataset.}
\label{table:cross_corr_dataset_OAI}
\begin{tabular}{ | p{3cm} | p{3cm} |  } 
  \hline
  Method & Left$+$right loss \\ 
  \hline
  Template matching & 0.221 \\ 
  \hline
  Neural method & 0.159  \\ 
  \hline
  Neural method + Sharpening step & 0.141  \\
  \hline
\end{tabular}
\end{table}

\begin{table}[!htbp]
\small
\centering
\caption{The sides combined average values over $400$ test images in the MOST dataset.}
\label{table:cross_corr_dataset_MOST}
\begin{tabular}{ | p{3cm} | p{3cm} |  } 
  \hline
  Method & Left$+$right loss \\ 
  \hline
  Template matching & 0.249 \\ 
  \hline
  Neural method & 0.224  \\ 
  \hline
  Neural method + Sharpening step & 0.157  \\
  \hline
\end{tabular}
\end{table}

\section{Conclusion}
In this paper, we considered detecting knee joint areas from bilateral knee X-ray images. For the annotations, we used just one side of a single bilateral image where we manually selected the templates. Otherwise, the training data in the teaching of the neural method was bilateral X-ray knee images without any labels. In the experiments it seemed that the neural network based method can give detections which are more uniform than detections from a classical version of the template matching. Overall, since the knee joint area varies quite rigidly between different knee images and there are no big deformations due to the knee anatomy across the images, it seems that in the knee joint area detection, it is not necessary to have many manually annotated images, one template image may be enough, provided the task-dependent energy function is properly chosen.  

\section*{Acknowledgments}
This work was done in a project funded by the Research Unit of Mathematical Sciences and the Research Unit of Medical Imaging, Physics and Technology at the University of Oulu, Finland.

\section*{Appendix: Grid initialized search algorithm}\label{appendix:grid}
In this section, we consider a grid initialized search algorithm to minimize $f^{\pmb{T}}(u,\pmb{v})$ in \eqref{eq:f}. We consider the template $\pmb{T}$  and the input halves as in Fig.\ \ref{fig:template_T}. Let $s\in\{left,right\}$. If $\pmb{\theta}^s=D(\pmb{v}^s),$ then $\pmb{v}^s$ is an unconstrained vector and $\pmb{\theta}^s$ describes the scale, translation and rotation. We form a grid  for $\pmb{\theta}^s$ and use the corresponding $\pmb{v}^s=D^{-1}(\pmb{\theta}^s)$ as an initial point for the unconstrained optimization problem. Here $\pmb{\theta}^s_1$ determines the scale of the patch we try to match to the template, $(\pmb{\theta}_2^s,\pmb{\theta}_3^s)$ determines the center of the patch. We set $\pmb{\theta}^s_4=0$ so there is no rotation in the initialization. We consider a few choices for the scale and for each scale we consider a dense enough grid for the translations such that adjacent rectangles are close enough. If the scale is large, i.e.\ $\pmb{\theta}^s_1\approx1$, then there is no need for so many translation grid points compared to when the scale is small, i.e.\ $\pmb{\theta}^s_1\approx0$.

First, we consider scales. We split the interval $[\alpha_0,\alpha_0+\beta_0]$, which is the range of \eqref{eq:scale_choice}, into $N_s$ number of equal distant points. Then we get $s_1=\alpha_0,\dots,s_{N_s}=\alpha_0+\beta_0$. How to initialize $d_j:=(\pmb{v}^s_j)_1$ such that we get $s_j$? Now
\begin{align*}
s_j&=\alpha_0+\beta_0\left(\frac{1}{2}\left(1+\tanh(d_j)\right)\right) \\
\Leftrightarrow d_j&=\tanh^{-1}\left(\frac{2(s_j-\alpha_0)}{\beta_0}-1\right).
\end{align*}
For translations, for a given scale $s_i$, we must have that $-s_i+\pmb{\theta}^s_2,s_i+\pmb{\theta}^s_2\in[-1,1]$ and $-s_i/f+\pmb{\theta}^s_3,s_i/f+\pmb{\theta}^s_3\in[-1,1]$ in order for the extracted patch to lie inside the input image. 

We get that we must have $s_i-1\leq\pmb{\theta}^s_2\leq 1-s_i$. We split the interval $[s_i-1,1-s_i]$ into $N_\theta$ equally spaced points $\theta_{13}^{i,j}$, $j=1,\dots,N_\theta$, such that $\theta_{13}^{i,1}=s_i-1,\dotsc,\theta_{13}^{i,N_\theta}=1-s_i$ and $\Delta^1$ denotes the distance between two consecutive grid points. How to choose the number of points $N_\theta$? We have $\theta_{13}^{i,j}=(s_i-1)+(j-1)\cdot\Delta^1$. Two $2 s_i$ sized  adjacent line segments have an overlap of size $2 s_i-\Delta^1$, when $\Delta^1<2 s_i$. If two adjacent line segments are required to have an  overlap whose length is at least $2 s_i-r 2 s_i$, where $0<r<1$, we get that $\Delta^1\leq r 2 s_i$. We set $\Delta^1=r 2 s_i$ with $r=0.25$ which leads to the choice $N_\theta=1+ceil(\frac{2-2s_i}{r 2 s_i})$. We note that $N_\theta$ increases as $s_i$ decreases which is what was desired.  

How to initialize $(\pmb{v}^s_{i,j})_2$ such that $\theta_{13}=(\pmb{\theta}_{i,j}^s)_2=\theta_{13}^{i,j}$? From \eqref{eq:tr_choice},
\begin{align*} 
\theta_{13}^{i,j}&=2(1-s_i)a_{13}^{i,j}-1+s_i \\
\Leftrightarrow a_{13}^{i,j}&=\frac{\theta_{13}^{i,j}+1-s_i}{2(1-s_i)} \\
\Leftrightarrow (\pmb{v}_{i,j}^s)_2&=\tanh^{-1}\left(\frac{\theta_{13}^{i,j}+1-s_i}{1-s_i}-1\right)
\end{align*}
and analogously for $\pmb{v}_3$. 

We finally consider the combination of the sides initializations.
If $[-s_i^{left}+\theta^{left,i,j}_{1,3},s^{left}_1+\theta^{left,i,j}_{1,3}]\subset[-1,1]$ is the projection of a left initialization rectangle into the horizontal axis,  we initialize the patch in the right side by using the same vertical initialization, but horizontally we test every initialization rectangle whose horizontal center is within the interval $[\theta^{left,i,j}_{1,3}-\frac{1}{3},\theta^{left,i,j}_{1,3}+\frac{1}{3}]$.

\bibliographystyle{abbrv}

\begin{thebibliography}{10}

\bibitem{Adarsh}
P.~Adarsh, P.~Rathi, and M.~Kumar.
\newblock Yolo v3-tiny: Object detection and recognition using one stage
  improved model.
\newblock In {\em 2020 6th International Conference on Advanced Computing and
  Communication Systems (ICACCS)}, pages 687--694, 2020.

\bibitem{Andrychowicz}
M.~Andrychowicz, M.~Denil, S.~G\'{o}mez, M.~W. Hoffman, D.~Pfau, T.~Schaul,
  B.~Shillingford, and N.~de~Freitas.
\newblock Learning to learn by gradient descent by gradient descent.
\newblock In D.~Lee, M.~Sugiyama, U.~Luxburg, I.~Guyon, and R.~Garnett,
  editors, {\em Advances in Neural Information Processing Systems}, volume~29.
  Curran Associates, Inc., 2016.

\bibitem{Antony2017AutomaticDO}
J.~Antony, K.~McGuinness, K.~Moran, and N.~E. O'Connor.
\newblock Automatic detection of knee joints and quantification of knee
  osteoarthritis severity using convolutional neural networks.
\newblock In {\em MLDM}, 2017.

\bibitem{Antony2020}
J.~Antony, K.~McGuinness, K.~Moran, and N.~E. O'Connor.
\newblock {\em Feature Learning to Automatically Assess Radiographic Knee
  Osteoarthritis Severity}, pages 9--93.
\newblock Springer International Publishing, Cham, 2020.

\bibitem{Antony2016QuantifyingRK}
J.~Antony, K.~McGuinness, N.~E. O'Connor, and K.~A. Moran.
\newblock Quantifying radiographic knee osteoarthritis severity using deep
  convolutional neural networks.
\newblock {\em 2016 23rd International Conference on Pattern Recognition
  (ICPR)}, pages 1195--1200, 2016.

\bibitem{CHEN201984}
P.~Chen, L.~Gao, X.~Shi, K.~Allen, and L.~Yang.
\newblock Fully automatic knee osteoarthritis severity grading using deep
  neural networks with a novel ordinal loss.
\newblock {\em Computerized Medical Imaging and Graphics}, 75:84--92, 2019.

\bibitem{Gatys}
L.~A. Gatys, A.~S. Ecker, and M.~Bethge.
\newblock Image style transfer using convolutional neural networks.
\newblock In {\em 2016 IEEE Conference on Computer Vision and Pattern
  Recognition (CVPR)}, pages 2414--2423, 2016.

\bibitem{GuLi}
H.~Gu, K.~Li, R.~J. Colglazier, J.~Yang, M.~Lebhar, J.~O'Donnell, W.~A.
  Jiranek, R.~C. Mather, R.~J. French, N.~Said, J.~Zhang, C.~Park, and M.~A.
  Mazurowski.
\newblock Automated grading of radiographic knee osteoarthritis severity
  combined with joint space narrowing.
\newblock {\em CoRR}, abs/2203.08914, 2022.

\bibitem{he_kaiming}
K.~He, X.~Zhang, S.~Ren, and J.~Sun.
\newblock Deep residual learning for image recognition.
\newblock In {\em 2016 IEEE Conference on Computer Vision and Pattern
  Recognition (CVPR)}, pages 770--778, 2016.

\bibitem{NIPS2015_33ceb07b}
M.~Jaderberg, K.~Simonyan, A.~Zisserman, and k.~kavukcuoglu.
\newblock Spatial transformer networks.
\newblock In C.~Cortes, N.~Lawrence, D.~Lee, M.~Sugiyama, and R.~Garnett,
  editors, {\em Advances in Neural Information Processing Systems}, volume~28.
  Curran Associates, Inc., 2015.

\bibitem{Kellgren_Lawrence}
J.~H. Kellgren and J.~S. Lawrence.
\newblock Radiological assessment of osteo-arthrosis.
\newblock {\em Ann Rheum Dis.}, 16(4):494--502, 1957.

\bibitem{Lewis95fastnormalized}
J.~Lewis.
\newblock Fast normalized cross-correlation.
\newblock {\em Ind. Light Magic}, 10, 2001.

\bibitem{Pierson}
E.~Pierson, D.~Cutler, J.~Leskovec, S.~Mullainathan, and Z.~Obermeyer.
\newblock An algorithmic approach to reducing unexplained pain disparities in
  underserved populations.
\newblock {\em Nature Medicine}, 27:136--140, 01 2021.

\bibitem{redmon_farhardi}
J.~Redmon and A.~Farhadi.
\newblock Yolo9000: Better, faster, stronger.
\newblock In {\em 2017 IEEE Conference on Computer Vision and Pattern
  Recognition (CVPR)}, pages 6517--6525, 2017.

\bibitem{SHAMIR20091307}
L.~Shamir, S.~Ling, W.~Scott, M.~Hochberg, L.~Ferrucci, and I.~Goldberg.
\newblock Early detection of radiographic knee osteoarthritis using
  computer-aided analysis.
\newblock {\em Osteoarthritis and Cartilage}, 17(10):1307--1312, 2009.

\bibitem{tiulpin2017novel}
A.~Tiulpin, J.~Thevenot, E.~Rahtu, and S.~Saarakkala.
\newblock A novel method for automatic localization of joint area on knee plain
  radiographs.
\newblock In {\em Scandinavian Conference on Image Analysis}, pages 290--301.
  Springer, 2017.

\bibitem{Wang_Bi}
Y.~Wang, Z.~Bi, Y.~Xie, T.~Wu, X.~Zeng, S.~Chen, and D.~Zhou.
\newblock Learning from highly confident samples for automatic knee
  osteoarthritis severity assessment: Data from the osteoarthritis initiative.
\newblock {\em IEEE Journal of Biomedical and Health Informatics}, pages 1--1,
  2021.

\end{thebibliography}

\end{document}